\documentclass[conference]{IEEEtran}
\IEEEoverridecommandlockouts
\usepackage{cite}
\usepackage{amsmath,amssymb,amsfonts}
\usepackage{color}
\usepackage{algorithmic}
\usepackage{graphicx}
\usepackage{textcomp}
\usepackage{xcolor}
\usepackage{booktabs}
\usepackage{makecell}
\usepackage{multicol}
\usepackage{multirow}
\usepackage{url}
\usepackage{amsthm} 
\usepackage{algorithm}
\usepackage{algorithmic}
\usepackage{subfigure}
\usepackage{booktabs}
\def\BibTeX{{\rm B\kern-.05em{\sc i\kern-.025em b}\kern-.08em
    T\kern-.1667em\lower.7ex\hbox{E}\kern-.125emX}}
\begin{document}

\title{Towards Fairer and More Efficient Federated Learning via Multidimensional \\
Personalized Edge Models
%\thanks{This work was supported by National Key R\&D Program of China No. 2020AAA0108800, NSFC under Grant 62172326 and 62137002, the MOE Innovation Research Team No. IRT\_17R86, China University Innovation Fund NO. 2021FNA04003, and the Project of China Knowledge Centre for Engineering Science and Technology.}
}

\author{\IEEEauthorblockN{Yingchun Wang}
\IEEEauthorblockA{\textit{School of Comp. Sci.\&Tech.} \\
\textit{Xi'an Jiaotong University}\\
Xi'an, Shaanxi, China\\
20116342r@connect.polyu.hk}
\and
\IEEEauthorblockN{Jingcai Guo}
\IEEEauthorblockA{\textit{Department of Computing} \\
\textit{The Hong Kong Polytechnic University}\\
Hong Kong SAR, China \\
jc-jingcai.guo@polyu.edu.hk}
\and
\IEEEauthorblockN{Jie Zhang}
\IEEEauthorblockA{\textit{Department of Computing} \\
\textit{The Hong Kong Polytechnic University}\\
Hong Kong SAR, China \\
18104473r@connect.polyu.hk}
\and
\IEEEauthorblockN{Song Guo}
\IEEEauthorblockA{\textit{Department of Computing} \\
\textit{The Hong Kong Polytechnic University}\\
Hong Kong SAR, China \\
song.guo@polyu.edu.hk}
\and
\IEEEauthorblockN{Weizhan Zhang}
\IEEEauthorblockA{\textit{School of Comp. Sci.\&Tech.} \\
\textit{Xi'an Jiaotong University}\\
Xi'an, Shaanxi, China\\
zhangwzh@xjtu.edu.cn}
\and
\IEEEauthorblockN{Qinghua Zheng}
\IEEEauthorblockA{\textit{School of Comp. Sci.\&Tech.} \\
\textit{Xi'an Jiaotong University}\\
Xi'an, Shaanxi, China\\
qhzheng@mail.xjtu.edu.cn}
}

\maketitle

\begin{abstract}
Federated learning (FL) is an emerging technique that trains massive and geographically distributed edge data while maintaining privacy. 
However, FL has inherent challenges in terms of fairness and computational efficiency due to the rising heterogeneity of edges, and thus usually results in sub-optimal performance in recent state-of-the-art (SOTA) solutions. 
In this paper, we propose a Customized Federated Learning (CFL) system to eliminate FL heterogeneity from multiple dimensions. Specifically, CFL tailors personalized models from the specially designed global model for each client jointly guided by an online trained model-search helper and a novel aggregation algorithm. 
Extensive experiments demonstrate that CFL has full-stack advantages for both FL training and edge reasoning and significantly improves the SOTA performance w.r.t. model accuracy (up to $7.2$\% in the non-heterogeneous environment and up to $21.8$\% in the heterogeneous environment), efficiency, and FL fairness.
\end{abstract}


\section{Introduction}
\label{sec:intro}
Machine deep learning has made tremendous success in the past few years across multiple real-world applications~\cite{deng2009imagenet,batista2004study,lin2011machine,guo2020novel,lv2014traffic,nguyen2015topic,guo2020dual,lian2018xdeepfm,he2016deep,simonyan2014very,guo2019adaptive,yang2015facial,shone2018deep,mahapatra2021medical,guo2023graph,liu2022towards,guo2021conservative,lu2023decomposed,wang2023data,liu2023zsl,guo2023application,huo2023offline,zhou2021device,guo2021learning,wang2022exploring,ma2019position,guo2016improved,huo2022procc,zhou2022cadm,guo2019ams,wang2022efficient,guo2019ee,zhou2021octo,guo2022fed}. In recent years, with the advances in the computing capability of edge devices, more and more machine learning models are usually deployed locally to directly perform training or inference. At the same time, the explosive growth of end-user data can have a high potential to bring about tangible benefits for various applications, i.e., user verification~\cite{face,lip}, self-driving~\cite{drive}, human activity recognition~\cite{human}, medical health monitoring~\cite{health}, and so on. 
However, the data resources are usually geographically distributed across different edge devices, this is then challenging to train a deep model collaboratively on both privacy preservation and transmission overhead reduction. Worse still, the ``isolated data islands'' could be seriously heterogeneous in terms of data distribution, quality, and quantity. 
Therefore, how to efficiently excavate the rich knowledge hidden behind these distributed data and train an accurate cooperative model in a privacy-protection way has become a crucial research topic.\par
To address the above challenges, federated learning (FL) has emerged as a promising paradigm of privacy-preserving distributed machine learning framework~\cite{FedAvg,tang2022personalized}. FL enables edge devices to train a model locally based on its own dataset and communicates with other models in the server for aggregating a more generalized global model, without collecting and sharing any privacy-sensitive information from users. 
However, existing FL methods mostly use a unified global model for all participants and ignore the potential hardware and data heterogeneities between them, i.e., diverse hardware specifications, different network conditions, highly biased data, and inconsistent data qualities. \par
In recent years, there have been some works focusing on eliminating the FL heterogeneity and obtained some promising results~\cite{1,2,3,4}.
However, most of them suffer from sub-optimization in both model performance and training efficiency, for only addressing the above problems in a single-dimensional perspective, i.e., data heterogeneity. 
Such a scheme may have three limitations: 1) the computation and transmission stragglers among different clients usually lead to extremely low training efficiency; 
2) the communication overhead of exchanging the updates of full models may cause excessive transmission delays, especially for large deep neural networks; 
and 3) the heterogeneities can lead to biased training across clients and introduce significant performance unfairness. 
%
\iffalse
\begin{figure*}[ht]
	\centering
	\includegraphics[width=0.8\textwidth]{fig1.jpg}
	\caption{Illustrations of three main heterogeneous challenges in federated learning scenarios} 
	\label{fig1}
\end{figure*}
\fi
%
In this paper, we propose a novel Multidimensional Customized Federated Learning (CFL) system to achieve a fairer and more efficient FL. 
The main idea of CFL is to strengthen the identity between the FL participants by minimizing the impact of multi-dimensional heterogeneities on FL. 
Specifically, a specially designed data-quality aware model with a tiny reinforcement learning (RL) module is regarded as the common model in CFL. 
Then, CFL tailors a personalized model from the common one for each client according to the corresponding device's hardware specification and data condition. 
The model tailoring is achieved by an online trained model-search helper which can output the submodel, i.e., network sub-structure, with the best accuracy and less latency for the current client. 
Moreover, to better aggregate information from asymmetric personalized models, we propose a novel parameter aggregation algorithm via module scaling and alignment in each FL round.
Therefore, the personalized models can be sampled dynamically from the common model with the data-aware RL module, and be trained further in an FL paradigm with the new scaling aggregation method. 
Extensive experiments on CIFAR-10 and MNIST datasets (processed as mixed quality) demonstrate that the proposed CFL can achieve better task accuracy, fairer model performance, and higher training acceleration against representative FL methods.
\begin{itemize}
\item We propose a new FL paradigm that tailors customized submodels for heterogeneous edge workers, which obviously reduces the time differences of local training on different devices and greatly speeds up the federated training stage.
\item We design a novel search helper to select customized models for different workers and a new model aggregation algorithm to aggregate updates with different architectures. Both proposals enable the use of FL in essentially heterogeneous edge computing.
\item We improve the parent model to be data quality-aware by adding an RL module and training it on special process datasets with different data qualities. It turns out that data quality-aware FL models perform better on real-world applications in edge computing scenarios than traditional FL paradigms.
\end{itemize}

\begin{figure*}[!ht]
	\centering
    \includegraphics[width = 0.85\textwidth]{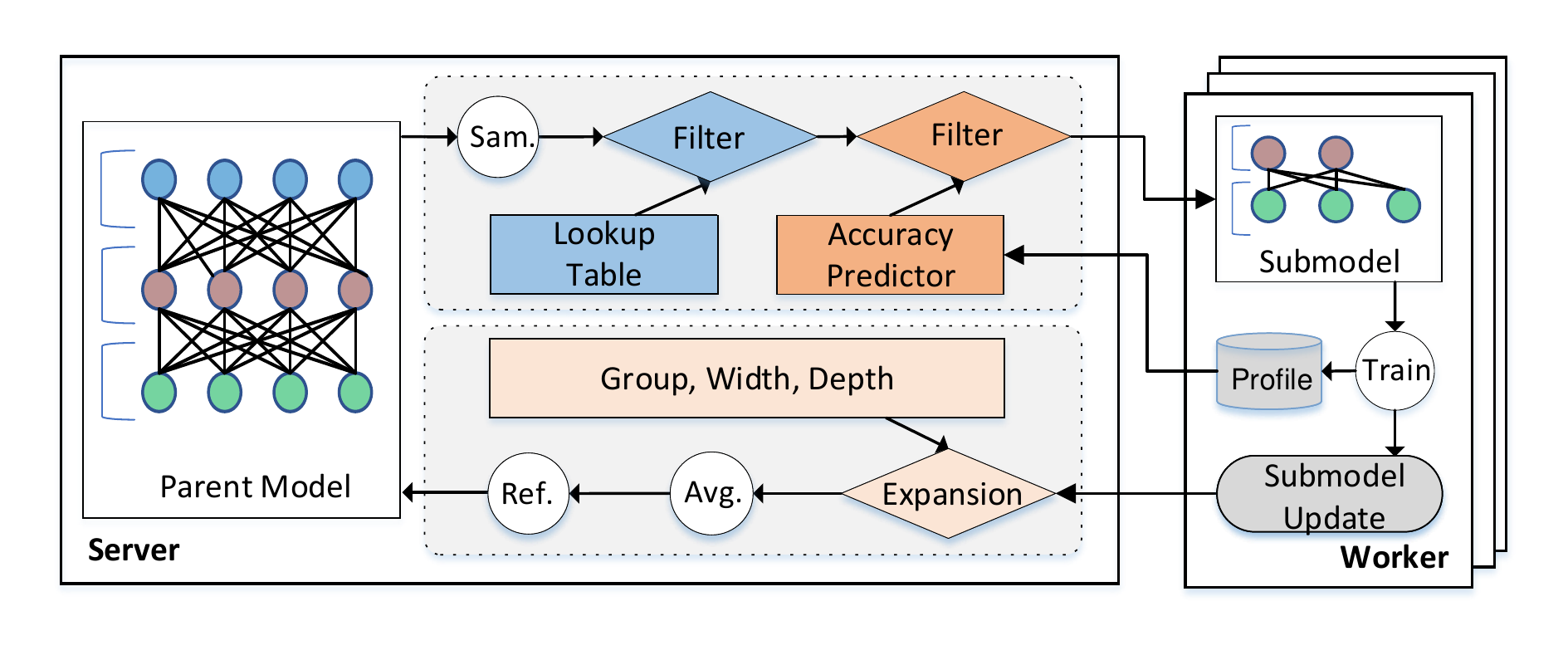}
	\caption{The overview of the CFL. The server part consists of three main components, i.e., the online accuracy predictor, the latency lookup table, and the expansion module, which work jointly to sample the optimal submodel for each client and update the global model. The blue lines on the parent model represent additional RL gating modules. Sam., Avg., and Ref. denote submodel sampling, parameter averaging, and the referenced module updating process respectively.} 
	\label{fig1}
\end{figure*}
\section{Related Works}\label{Section:RelatedWork}

\subsection{Model Compression} 
Model compression has been extensively studied to reduce the computation overhead and memory usage, so that neural models can be better deployed on resource-limited edge devices \cite{DC, wang2021survey}. It is essential to do a trade-off between the compressed rate and the accuracy reduction. Current popular methods include pruning \cite{Han2015Learning}~\cite{Anwar2015Structured}~\cite{Rueda2017Neuron}, quantization~\cite{Li2017Performance}, parameter sharing~\cite{Chen2015Compressing}, knowledge distillation, low-rank approximation and direct design of compact models, etc. 
% Besides, another similar method named Neural Architecture Search (NAS) is also widely considered to explore a suitable lightweight model for resource-limited devices.  NAS usually deploys a well-trained model on the server as the parent and trains a neural network \cite{ofa20} or RL agent \cite{tan2019mnasnet} to help to choose proper submodels from the parent model to adapt dynamic runtime resources on distributed platforms. 
We compare our work with several mainstream-related works including standard FL \cite{FedAvg}, MT-FL \cite{MTL}, and Model compression. The details are shown in Table~\ref{table1}.
  \begin{table*}[htbp]
  	\caption{Comparison of some works related to edge computing}
  	\label{table1}
  	\centering
  	\begin{tabular}{|p{1.6cm}<{\centering}|p{1.2cm}<{\centering}|p{0.6cm}<{\centering}|p{5cm}<{\centering}|p{1.2cm}<{\centering}|p{0.6cm}<{\centering}|p{3.5cm}<{\centering}|}
  		\hline
  		\textbf{Methods}            & \multicolumn{3}{c|}{\textbf{Heterogeneous in training}}                                                            & \multicolumn{3}{c|}{\textbf{Heterogeneous in inference}}                                                                             \\ \cline{2-7}
  		\multicolumn{1}{|l|}{}      & \textbf{Device}  & \textbf{Data}      & \textbf{Idea}                                           & \textbf{Device}  & \textbf{Data}      & \textbf{Idea}                                                             \\ \hline
  		Standard FL \cite{FedAvg}                & N                  & N                  & All clients train the same model                        & N                  & N                  & All clients use the same trained model                                    \\ \hline
  		\multirow{2}{*}{MT-FL \cite{MTL} }     & \multirow{2}{*}{N} & \multirow{2}{*}{Y} & Same structure for collaboration                        & \multirow{2}{*}{N} & \multirow{2}{*}{Y} & \multicolumn{1}{l|}{\multirow{2}{*}{Each client uses its specific model}} \\ \cline{4-4}
  		&                    &                    & Different parameters for data heterogeneity             &                    &                    & \multicolumn{1}{l|}{}                                                     \\ \hline
  		Model compression  \cite{Chen2015Compressing}            & N                  & N                  & Training a large model in cloud                         & Y                  & N                  & \multicolumn{1}{p{3.5cm}|}{Compress the large model into a small model}            \\ \hline
  		\multirow{2}{*}{CFL (ours)} & \multirow{2}{*}{Y} & \multirow{2}{*}{Y} & Different structures for both types of heterogeneity & \multirow{2}{*}{Y} & \multirow{2}{*}{Y} & \multirow{2}{*}{Each client uses its specific model}                      \\ \cline{4-4}
  		&                    &                    & The same parameters for collaboration                   &                    &                    &                                                                           \\ \hline
  	\end{tabular}  	
  \end{table*}
  
\subsection{Federated Learning}
Applying big data analysis and artificial intelligence (AI) in practical scenarios is not easy since high-quality datasets are rare or difficult to access. To deal with the data island problem and utilize the distributed data from different mobile devices, Google has proposed federated learning (FL)~\cite{FL1,FedAvg} in 2016 to train machine learning models over decentralized devices. Such a paradigm aims to mine distributed datasets without sharing privacy-sensitive data, and thus is deemed to greatly help launch AI in more areas. FL requires a set of workers to cooperate with the coordinator server, which assigns the model parameters to each worker to train the local model and collects the updated parameters to accumulate to the shared model repeatedly. Current focuses of FL research include user privacy preservation \cite{privacy}, incentive scheme to promote collaboration between multiple users \cite{incentive}, communication overhead \cite{commu} and fairness optimization ~\cite{fair}.

\subsection{Heterogeneity in Data and Devices}
Data heterogeneity here is defined as ``the datasets from different devices are often statistically deficient (non-IID), e.g., of different label distribution, dataset size and sample noise level, etc". Device heterogeneity refers to the fact that different devices participating in FL tasks are highly likely to possess quite different levels of hardware specification and environmental conditions, e.g., different CPU spec, memory size, and network bandwidth. Heterogeneity in distributed scenarios leads to inevitable waiting latency for parameter aggregation as well as the variance of model accuracy among different devices. Lots of works have been proposed to deal with such problems. For example, the vertical FL is proposed to deal with feature heterogeneity \cite{survey}.

Several approaches to personalized federated learning \cite{pfl1,pfl2,pfl3} have been proposed to address the heterogeneity problem in FL. For example, Zhao et al. \cite{pflzhao} used the method of data enhancement to reduce the statistical heterogeneity between customer data sets and strengthen the training of the whole play model by sharing some balanced global data among drifting clients. 
Li et al. \cite{pflli} proposed the FedMD method to train a global model with the assistance of a public dataset and allow each client to fine-tune with its own private dataset. 
However, most of the above methods only consider the heterogeneity of data distribution between clients during training for personalized design, ignoring the common heterogeneous problems such as hardware and data quality. 
How to incorporate diverse heterogeneities remains a serious challenge in FL.

\section{Methodology}
\label{sec:Methodology}
\subsection{An Overview of CFL}\label{Subsection:Overview}

The pipeline of CFL is illustrated in Figure.~\ref{fig1}. Specifically, CFL can be divided into the interwoven local training and global updating process that follows a three-part paradigm: \textbf{submodels sampling} (on the server), \textbf{local training} (on the client), and \textbf{model updating} (on the server). 

First, in the submodels sampling stage, the server samples a personalized network sub-structure for each FL worker according to its hardware specification and data quality, wherein the hardware specification is defined by the device model, and the data quality is quantized to five different quality levels according to the Gaussian Blur. 
To be specific, the submodels are first sampled from the global (parent) model using ``Genetic Algorithm'' on both model depth and width dimensions.
After that, the selected submodel will be further filtered through a search helper composed of an online-trained accuracy predictor and an offline latency lookup table~\cite{ofa20}. 

Next, the sampled submodels are assigned to different clients and be trained with heterogeneous local datasets separately. At the last epoch of local training, the test accuracy of each submodel is collected as the training profile, which is then be uploaded along with the model gradients.\par

Last, there are two model updating processes parallelly conducted in the server. 
The first update is about the \textbf{global model}, which aggregates all the uploaded local updates from clients by the specially designed aggregation algorithm. 
The second update is about the \textbf{accuracy predictor}, which judges the accuracy that the current submodel can achieve based on the hardware and data conditions reported by each client. \par

\subsection{Personalized Models For Device Heterogeneity}\label{Subsection:Customized submodels}
In this subsection, we mainly focus on problems introduced by device heterogeneity in traditional federated learning during the training stage. Specifically, there are two main problems. The first comes from the selection of customized submodels, one of the filters of which needs an accurate and less training overhead accuracy predictor. And another challenge would be the above-mentioned submodels aggregation. These two issues are respectively discussed in the following two subsections below.
\subsubsection{submodels Sampling}\label{Subsubsection:Customization}
The submodel sampling process is based on two sequential steps, including submodel searching and filtering. To be specific, submodels are firstly randomly generated using genetic algorithms in a two-dimensional-limited search space. The details can be found in Algorithm \ref{alg1}.
\begin{algorithm}[htbp]
	\caption{submodels Selection}
	\label{alg1}
	\begin{algorithmic}
		\STATE {\bfseries Input:} Accuracy predictor $f_{t}$, computational latency table $g$, parent model $\omega_t$, search times $S$, computational latency bound $l_{k}$, hardware profile $p_{k}$, data quality $q_{k}$, number of workers $K$ 
		\STATE {\bfseries Output:} Customized models $\omega_{k}^t$ for each worker $k$
		\STATE Initialize $acc_k=0, \forall\ k=1,\ldots,K$;
		\FOR{$i=0, ..., S$}
    		\FOR{$k=0, \ldots ,K$}
        		\STATE Select a submodel $\omega$ from $\omega_t$ with the bounded latency on worker $k$ as $g(\omega, p_{k})<l_{k}$;
        		\IF{$f_{t}(\omega,q_{k})>acc_k$}
            		\STATE $acc_k = f_{t}(\omega, q_{k})$;
            		\STATE $\omega_{k}^t=\omega$;
        		\ENDIF
    		\ENDFOR
    	\ENDFOR
	\end{algorithmic}
\end{algorithm}
After that, these generating results will be further filtered through a search helper composed of an online-trained accuracy predictor and an offline latency lookup table~\cite{ofa20}, 
wherein, the accuracy predictor is essentially a four-layer linear classifier and is dynamically trained in the first several FL rounds. %
Its training datasets are the above-mentioned training profiles, which are formed by using the submodel structure information and data quality as the data sample, and the test accuracy as the data label.
Since the accuracy predictor is a simple linear classifier and the number of clients in FL is large, the data samples and labels collected from one or two CFL rounds would be sufficient enough to let the accuracy predictor converge or reach a satisfying prediction accuracy. Once the training of the accuracy predictor starts to converge, or its test accuracy reaches a predefined threshold, its training can be stopped to stabilize submodels as well as reduce overhead. 

In subsequent rounds, the predicted accuracy, together with the latency table, are used for the selection process. A complete training process in one CFL round of the accuracy predictor is shown in Algorithm~\ref{alg3}. 
\begin{algorithm}[!t]
	\caption{Training Accuracy Predictor}
	\label{alg3}
	\begin{algorithmic}
	    \STATE {\bfseries Input:} The accuracy predictor $f$ 
		\STATE {\bfseries Output:} $f$ in each FL round
		\FOR{$t=1$ to $T$}
    		\FOR{$k=1$ to $K$}
    		    \STATE Collect $q_{k}$, $\omega_{k}^t$, $acc_{k}^t$ from the worker $k$;
        		\STATE Construct sample $x_k=(q_{k}, \omega_{k}^t)$, $y_k=acc_{k}^t$;
    		\ENDFOR
    		 \STATE Use all $K$ samples $(x,y)$ to update the accuracy predictor $f_t$ for one epoch;
		\ENDFOR
	\end{algorithmic}
\end{algorithm}

\subsubsection{submodels Aggregation}\label{Subsubsection:Aggregation}
To solve the challenges of the aggregation of structurally misaligned personalized submodels, we proposed a novel aggregation algorithm that expands and aligns all submodels before the actual aggregation. Same with model sampling, the expansion of submodels is also limited in depth and width dimension. The details are as follows.
%
\iffalse
\begin{figure}[tp]
	\centering
	\subfigure[Layer group in the parent model with residual architecture.]{
		\includegraphics[width=0.44\textwidth]{fig3.jpg} 
	}\\
	\subfigure[Layer group in a customized submodel during aggregation.]{
		\includegraphics[width=0.4\textwidth]{group.jpg} 
	}
	
	\caption{(a) illustrates the vanilla group setting on the parent model with residual architecture, where a group represents a set of convolutional layers with some of the same properties, i.e, a residual block in ResNet. And b shows the layer grouping in customized models before and after aggregation of CFL respectively.}
	\label{group}
\end{figure}	
\fi
%
\begin{figure}[htbp]
		\centering
		\includegraphics[width = 0.38\textwidth]{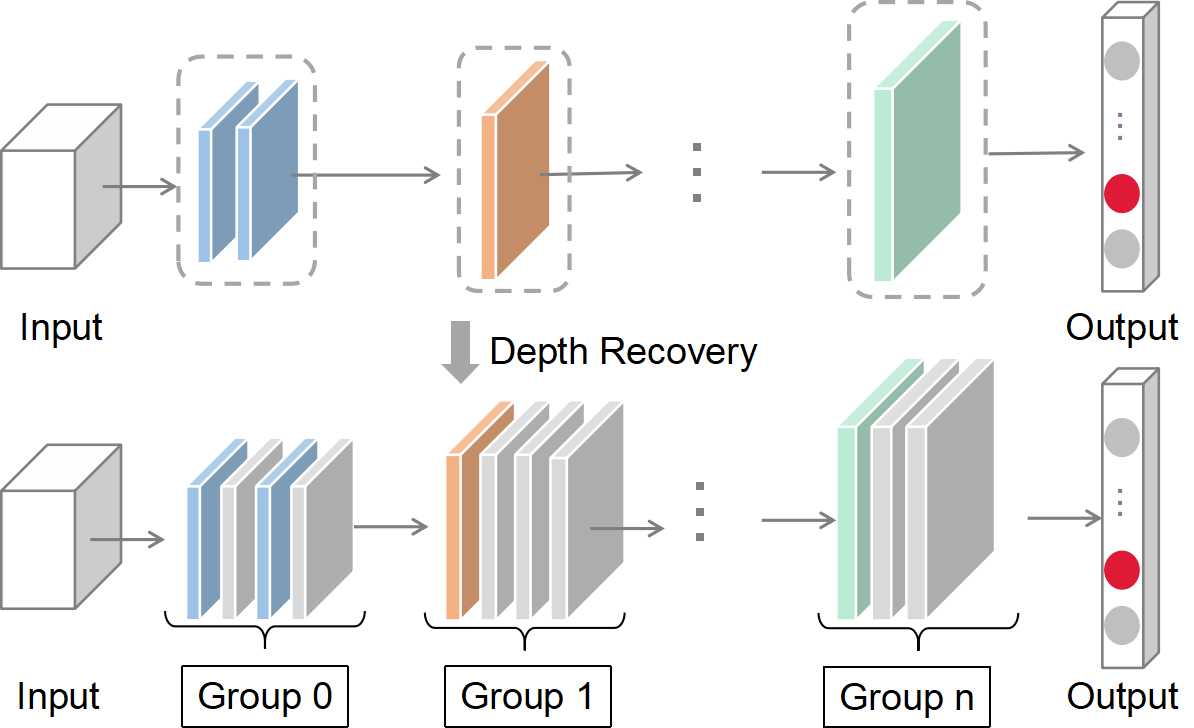} 
		\caption{Depth expansion of submodel.} 
		\label{depth} 
	\end{figure}

\begin{figure}[htbp]
		\centering
		\includegraphics[width = 0.38\textwidth]{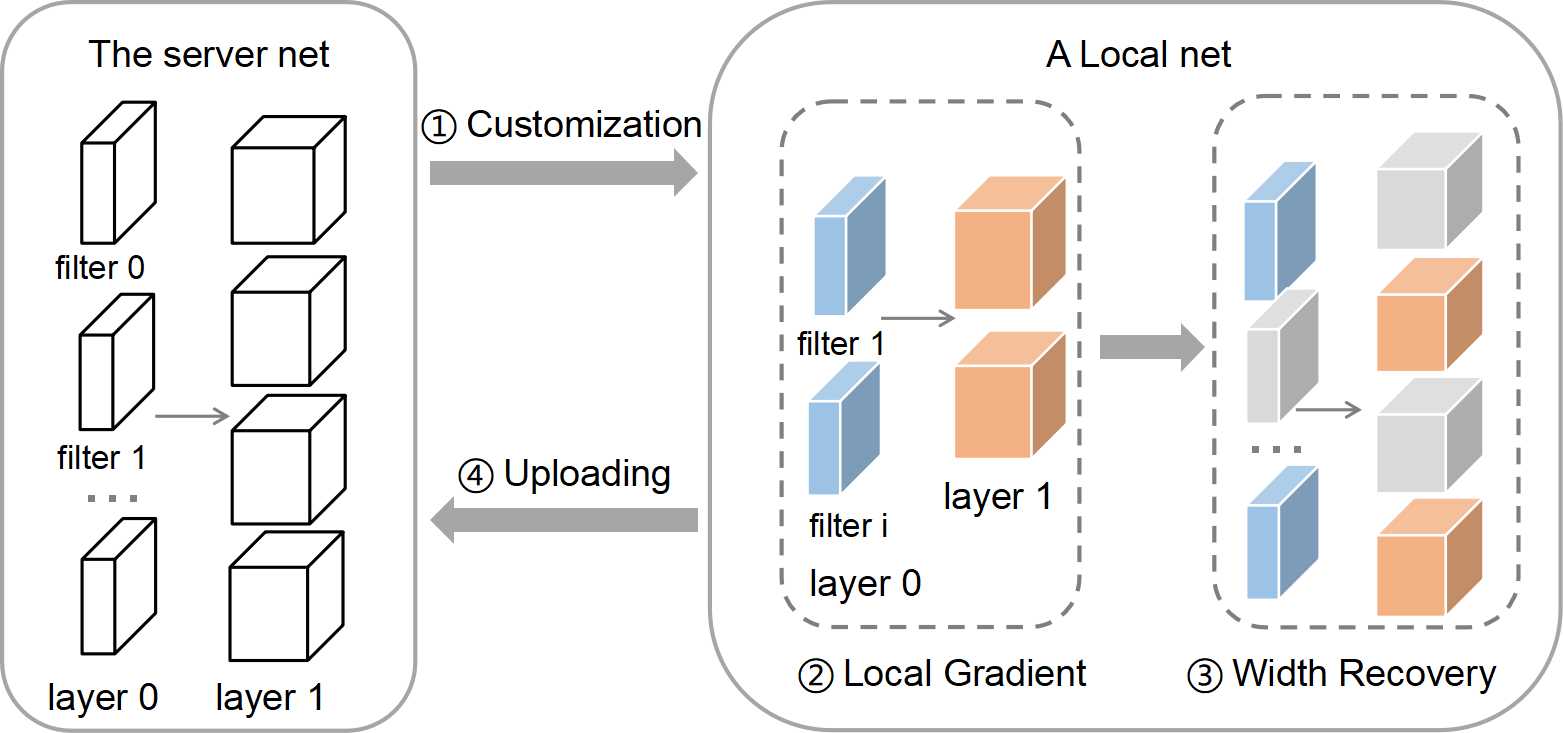} 
		\caption{Width expansion of customized models. The left rectangle represents the width of layers in the global model. The right rectangle represents the width of the layers in the submodels.} 
		\label{width}
	\end{figure}
\begin{itemize}
	
	\item Layer group: Since a residual model architecture is used in the parent model, different parameter settings and activation functions are used in different residual blocks. Before the actual model depth and layer width alignment, a layer grouping process is necessary to maintain the same parameter distribution as the parent model before and after expanding operation. Specifically, all the convolution layers except the first one of the submodels are divided into different groups according to the residual settings of the server model to achieve a group-level alignment. 

	\item Width expansion: Since the channels of each layer of the submodels are randomly selected from the parent model according to the limitations of the accuracy predictor and latency lookup table. In the vanilla FL aggregation stage, it is the parameters of the channels at the same location in different submodels participate in one operation. To face the challenge of the disordered channels of various submodels that are scrambled during the sampling process, as shown in Fig.~\ref{width}, all channels in the submodels are first sorted in the original order to keep the consistency of the parent model structure. And then, after the sorting process,  an expansion operation is performed for each layer of the submodels, if its current layer width is smaller than the width of the large model, all $0$ channels will be added to fill the current layer to its original width in the parent model. Thus far, all submodels have achieved width alignment.

	\item Depth expansion: Depth alignment is performed group-wise and can be done only after the group alignment. As shown in Fig.~\ref{depth}, for those groups with fewer layers in submodels than the parent model, they are padded with all $0$ layers to reach the layer number of the parent standard. Furthermore, the width and kernel size in these all-zero layers are the same as the width of the corresponding layer of the parent model.  
\end{itemize}	

After the model alignment, the federated average algorithm could be performed to aggregate all of the uploaded local grads and update the global model. The detailed working flow of the whole alignment and aggregation operation is given in Algorithm \ref{alg3}. 
\begin{algorithm}[htbp]
	\caption{submodel Alignment and Aggregation}
	\label{alg3}
	\begin{algorithmic}
		\STATE {\bfseries Input:} The number of workers $K$, submodel update $\Delta_k^t$ and data size $n_k$ for each worker $k$, and total data size $n$;
		\STATE {\bfseries Output:} Global model $\omega_t$;
		\FOR{$k=1$ to $K$ in parallel}
    		\STATE Group the layers of the update $\Delta_{k}^{t}$ by block;
    		\STATE Expand the width of the layers of $\Delta_{k}^{t}$;
            \STATE Extend the depth of $\Delta_{k}^{t}$;
		\ENDFOR
		\STATE Aggregate all updates $\Delta_{t} = \sum_{k=1}^K \frac{n_{k}}{n} \Delta_{k}^{t} $;
	\end{algorithmic}
\end{algorithm}
\subsection{RL-Based parent model For Data Heterogeneity}
After the training stage of FL, the optimized models (or submodels) would be deployed on edge devices in the real world.
And a machine learning model in the wild (e.g., a self-driving car) must be prepared to make sense of its surroundings in rare conditions that may not have been well-represented in its training set. However, the previous personalized FL only focuses on the heterogeneity of data distribution and ignores the changes in data quality in practice.

Previous work \cite{skip} has proved that to achieve the same prediction accuracy, heterogeneous data quality requires different network complexity. For example, a clear image may only require a smaller neural network to achieve the same accuracy as a blurred image.

Thus, to achieve fairer task performance across heterogeneous datasets, we enable the global model to be data-aware by an RL gating module to assign personalized submodels according to different data conditions. The RL function could dynamically select which layers of a convolutional neural network should be skipped during submodel sampling. 
Specifically, we first introduce layer-wise RL agents which are coded as a function from the feature activations to the probability distribution over the skipping action.
Note that to cope with non-differentiable data-aware model sampling decisions, we first warm up the global model and train it using a hybrid learning algorithm combining supervised learning and reinforcement learning~\cite{skip}.
It turns out that the RL modules in CFL not only speed up the FL edge training but also could accelerate the following inference stage by assigning more lightweight models to edge devices. In a word, CFL is a great full-stack FL system that could greatly reduce the computing overhead in both the training and inference stages. 

Last, the overall process of our method is summarized in Algorithm~\ref{alg:framework}.

\begin{algorithm}[htbp]
\caption{CFL: Customized Architecture Search based Federated Learning}
\label{alg:framework}
  \begin{algorithmic}
    \STATE {\bfseries Input:} Number of workers $K$, learning rate $\eta$
    \STATE {\bfseries Output:} local updates $\Delta_k^t$ and test accuracy of worker $k$ in communication round $t$
    \STATE {\bfseries Initialize:} parent model $\mathbf{\omega}_1$;
    \FOR{$t=1$ {\bfseries to} $T$}
      \STATE \underline{{\bfseries On server}:}
        \STATE \textit{Select} the submodel $\omega_k^t$ from $\omega_{t}$ for each worker $k$ by using the search helper;
        \STATE Send submodels $\omega_k^t$ to all workers ;
        \STATE Receive and \textit{aggregate} $\Delta_k^t$ of all workers to get $\Delta_t$; 
        \STATE Update the global model $\omega_{t+1} = \omega_{t} - \Delta_t$;
        \STATE Receive data and hardware profile and use them to update the search helper;
      \STATE \underline{{\bfseries On worker} k=1,\ldots,K:}
        \STATE Receive submodel $\omega_k^t$ from server;
		\FOR{epoch $e$ ranges from $1$ to $E$}
            \STATE Compute the stochastic gradient $\nabla l(\omega_{k,e-1}^t)$ from a random mini-batch;
    		\STATE Update submodel $\omega_{k,e}^t = \omega_{k, e-1}^t -\eta \nabla l(\omega_{k,e-1}^t)$;
		\ENDFOR
        \STATE Compute local update $\Delta_k^t = \omega_{k,E}^t - \omega_{k,0}^t$
        \STATE Send $\Delta_k^t$ to server;
        \STATE Send test accuracy and hardware specification profile to server;
    \ENDFOR
\end{algorithmic}
\end{algorithm}

\section{Experiment}
\label{sec:Result}

\subsection{Benchmark}
\textbf{Dataset}:
We use CIFAR-10 and MNIST datasets as the baseline, and a set of related datasets are extended from these raw datasets to emulate two different data heterogeneities, such as \textbf{data quality heterogeneity} and \textbf{data distribution heterogeneity}. 
For quality heterogeneity, the raw dataset is independently and identically (IID) divided into several batches and processed by Gaussian blurring and image sharpening with different levels. These batches are re-mixed to form new mixed-quality datasets.
For distribution heterogeneity, each dataset is randomly divided into 32 Non-IID subsets. The data class imbalance degree is set to 0.8 in this work, i.e., 80\% of each worker's local data belongs to the same class, and the remaining 20\% are evenly selected from the remaining categories. 
CIFAR-10 dataset is processed to simulate the data quality heterogeneity. It consists of 60000 32x32 color images in 10 classes, with 6000 images per class and 10000 images per batch. There are five batches of images for training and one batch of images for the test. To simulate data quality heterogeneity during practical inference, we use three different degrees of Gaussian blur and image sharpening on the CIFAR-10 dataset, one per batch, to produce datasets of different quality but the same distribution. To be specific, we divided the training set of CIFAR-10 into five groups: unprocessed, three degrees of Gaussian blur, and sharpening images respectively. Instead of using complex data quality metrics, we apply different variances of the added noise to represent the heterogeneity in data quality. 
MNIST dataset is used to generate data heterogeneity in both distribution and quality. As for the data quality heterogeneity, we divide the MNIST training set uniformly into five IID subsets and conduct the above-mentioned Gaussian blur or image sharpening per group, to produce datasets of different quality but the same distribution. As for the data distribution heterogeneity, we divide the whole MNIST dataset randomly into 32 Non-IID subsets to simulate the different data distribution between federated workers. The data class imbalance degree is set to 0.8 in this work, i.e., 80\% of each worker's local data belongs to the same class, and the remaining 20\% are evenly selected from the remaining categories. For the parent model, it is pre-trained on quality heterogeneous IID datasets, and then federally trained on quality heterogeneous and Non-IID datasets.

\textbf{Model}:
We use a once-for-all network \cite{ofa20} with layer-wise RL gate as the parent model, which is built on MobileNetV3 with elastic depth, width, and input size. All of the customized submodels in our experiments will be selected from the parent model. 

\subsection{The Comparison of CFL with FL SOTA}\label{subsec:cus and same}
\begin{figure}[ht]
	\centering
    \subfigure[Quality  heterogeneity]{
		\includegraphics[width=0.36\textwidth]{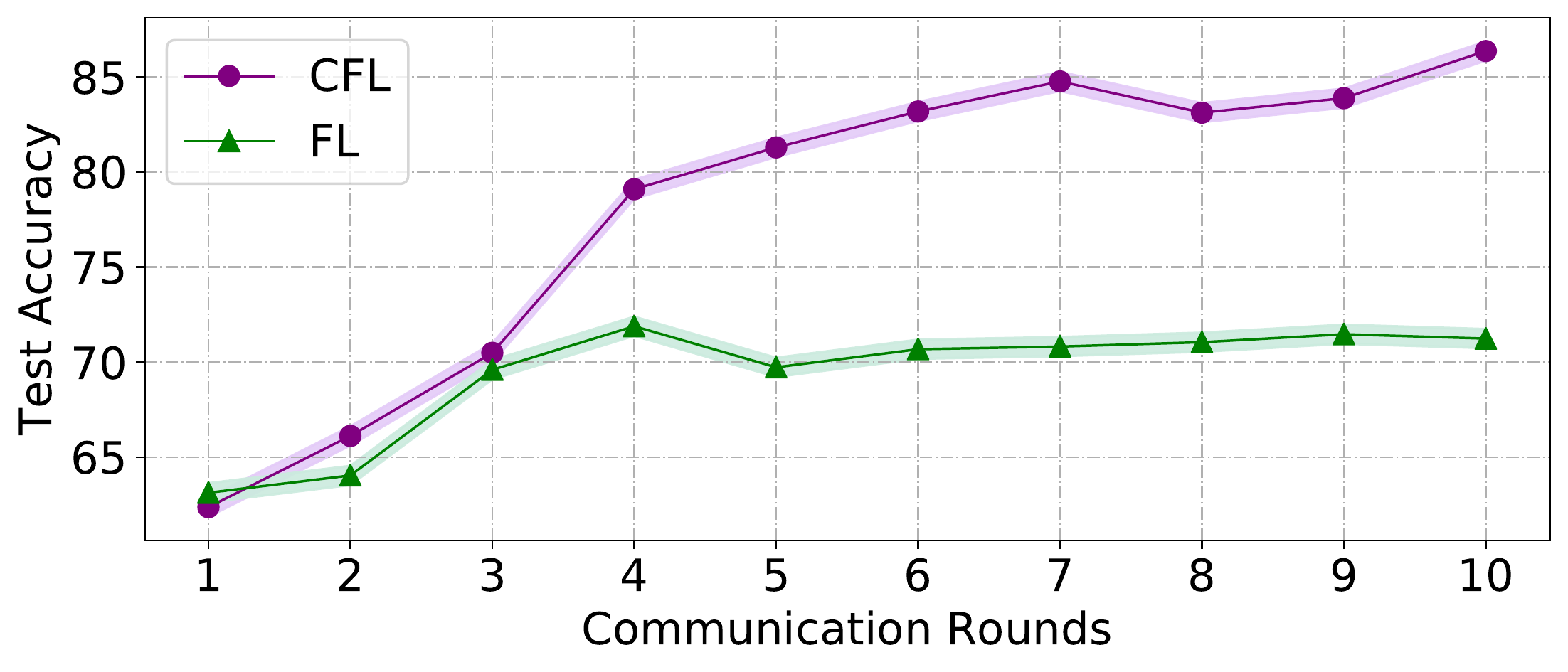} 
	}
	\subfigure[Distribution heterogeneity]{
		\includegraphics[width=0.36\textwidth]{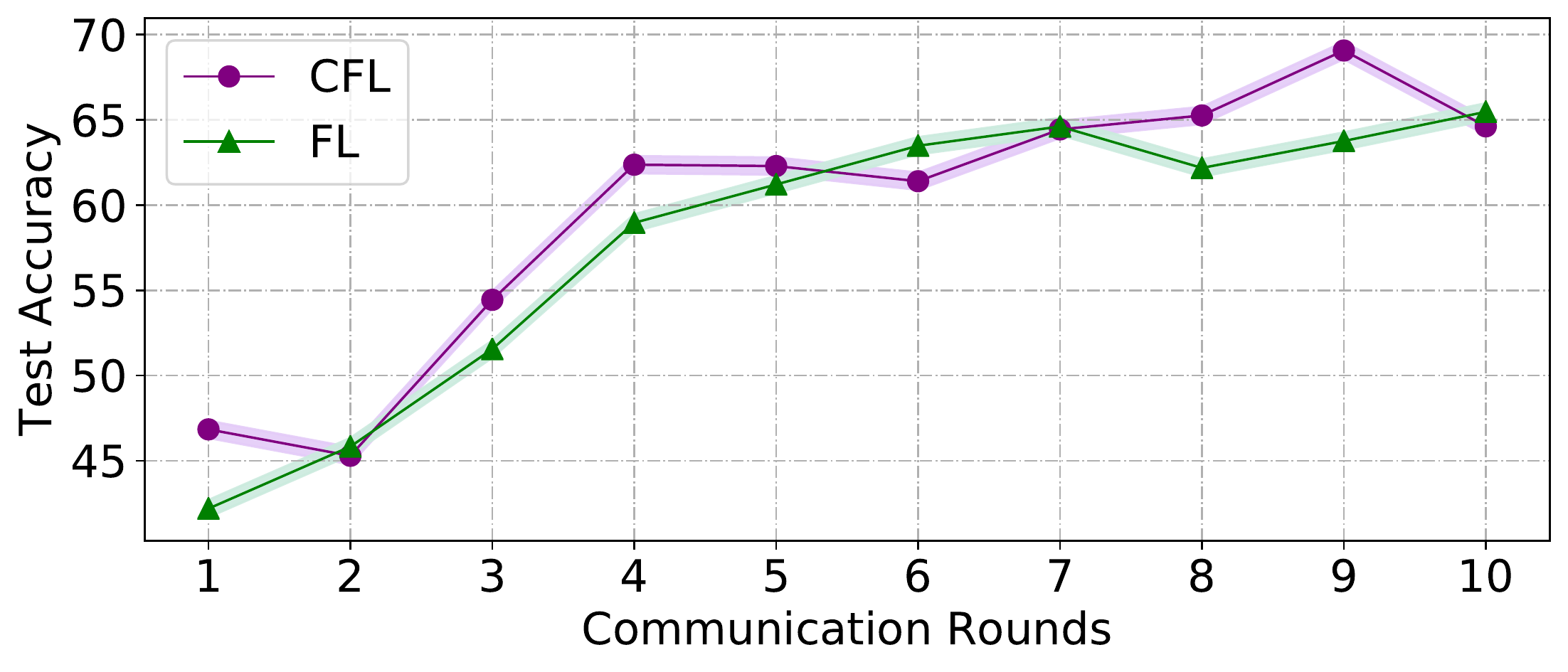} 
	}
	\caption{(a) Quality heterogeneity. (b) Distribution heterogeneity.}
	\label{subvsofa}
\end{figure}

\begin{figure}[t]
	\centering
	\includegraphics[width=0.36\textwidth]{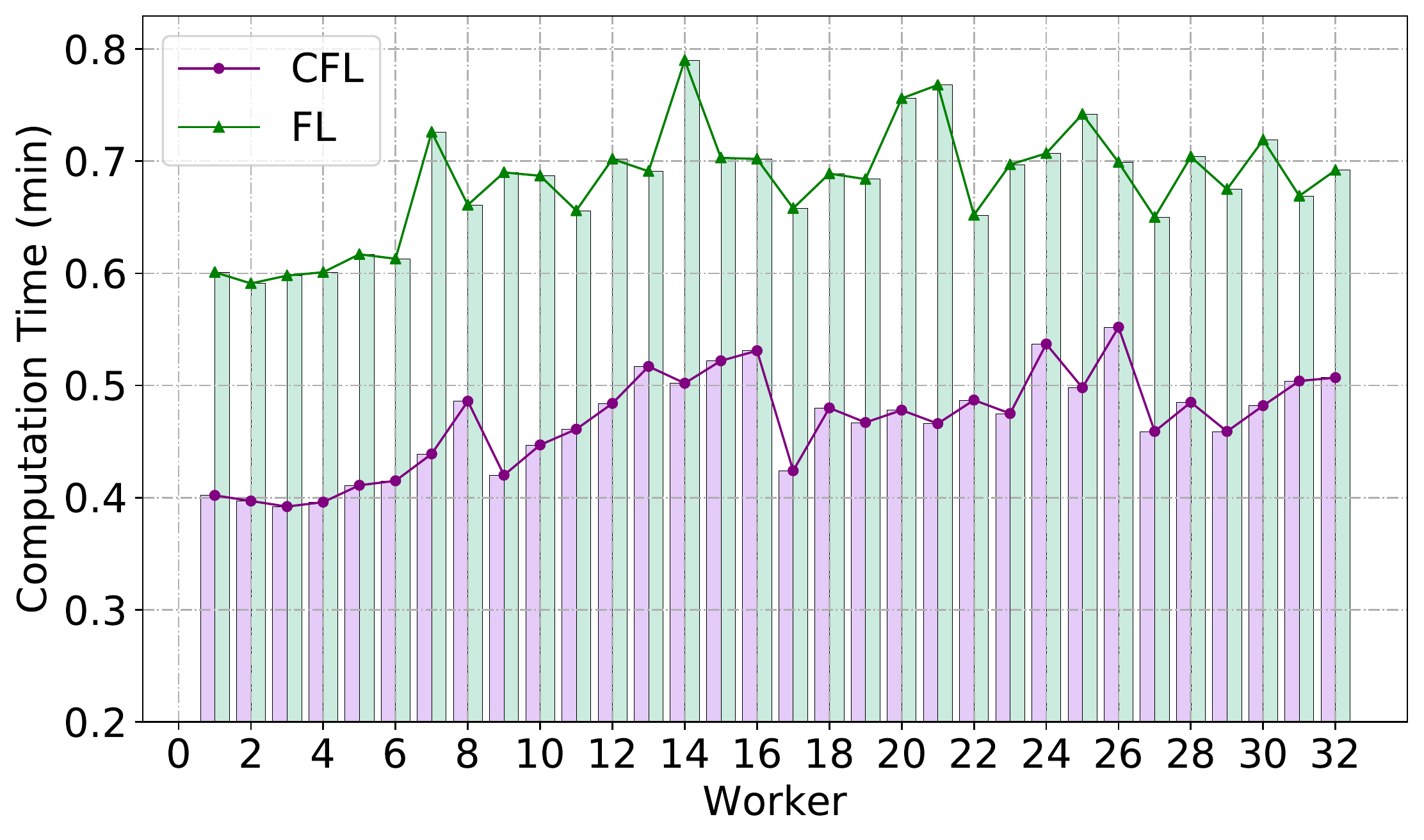}
	\caption{Time required for 200 iterations on 32 workers.}
	\label{time}
\end{figure}

In this section, we compare the performance of CFL using personalized models and FL SOTAs using one global model (abbreviated as FL in the following) on two different data heterogeneity settings with respect to data quality and distribution. The results are shown in Figure. \ref{subvsofa} (a) and Figure.\ref{subvsofa} (b), respectively. 
It is obvious that CFL performs significantly better than FL under both heterogeneous settings, especially when the data quality is heterogeneous.
Figure. \ref{time} shows the time required for the first 200 iterations over 32 clients of CFL and FL, respectively.
\par
The results demonstrate that CFL not only significantly improves training efficiency but also improves FL fairness because the training time difference between clients is significantly reduced. 

\subsection{Federated Learning vs. Independent Learning}\label{subsec:fl and ind}
The performance gain from CFL to local independent learning (IL) using customized models is demonstrated. The training experiments are conducted over two categories of datasets, i.e., MNIST with both heterogeneous quality and heterogeneous distribution, which are split into 32 subsets for each worker respectively. The result of the accuracy comparison between CFL and independent local training is shown in Table \ref{table3}, it is obvious that using CFL in model training for edge computing consistently outperforms independent training in both heterogeneous and non-heterogeneous data distributions. And, in the case of data heterogeneity, the advantages of CFL are further amplified.\par
\begin{table}[htbp]
\renewcommand\arraystretch{1.2}
\centering
	\caption{Comparison of test accuracy under two edge computing settings: CFL and Independent local training.}
	\label{table3}

	\begin{tabular}{l|cc|cc}
		\bottomrule
		& \multicolumn{2}{c|}{Non-heterogeneous Data}                                                                                             & \multicolumn{2}{c}{Heterogeneous Data}                                                                                                 \\ \hline
		scenario & \begin{tabular}[c]{@{}l@{}}CFL (\%)\end{tabular} & \begin{tabular}[c]{@{}l@{}}IL (\%)\end{tabular} & \begin{tabular}[c]{@{}l@{}}CFL (\%)\end{tabular} & \begin{tabular}[c]{@{}l@{}}IL (\%)\end{tabular} \\ \hline
		worker 0 & 82                                                                & 74.8                                                                & 80.5                                                              & 68.7                                                                \\ \hline
		worker 1 & 73.6                                                              & 73.4                                                                & 72.6                                                              & 50.8                                                                \\ \hline
		worker 2 & 86.1                                                              & 82.3                                                                & 85.6                                                              & 69                                                                  \\ \bottomrule
	\end{tabular}
\end{table}
\begin{figure}[tp]
	\centering
	\subfigure[Quality heterogeneity]{
		\includegraphics[width=0.36\textwidth]{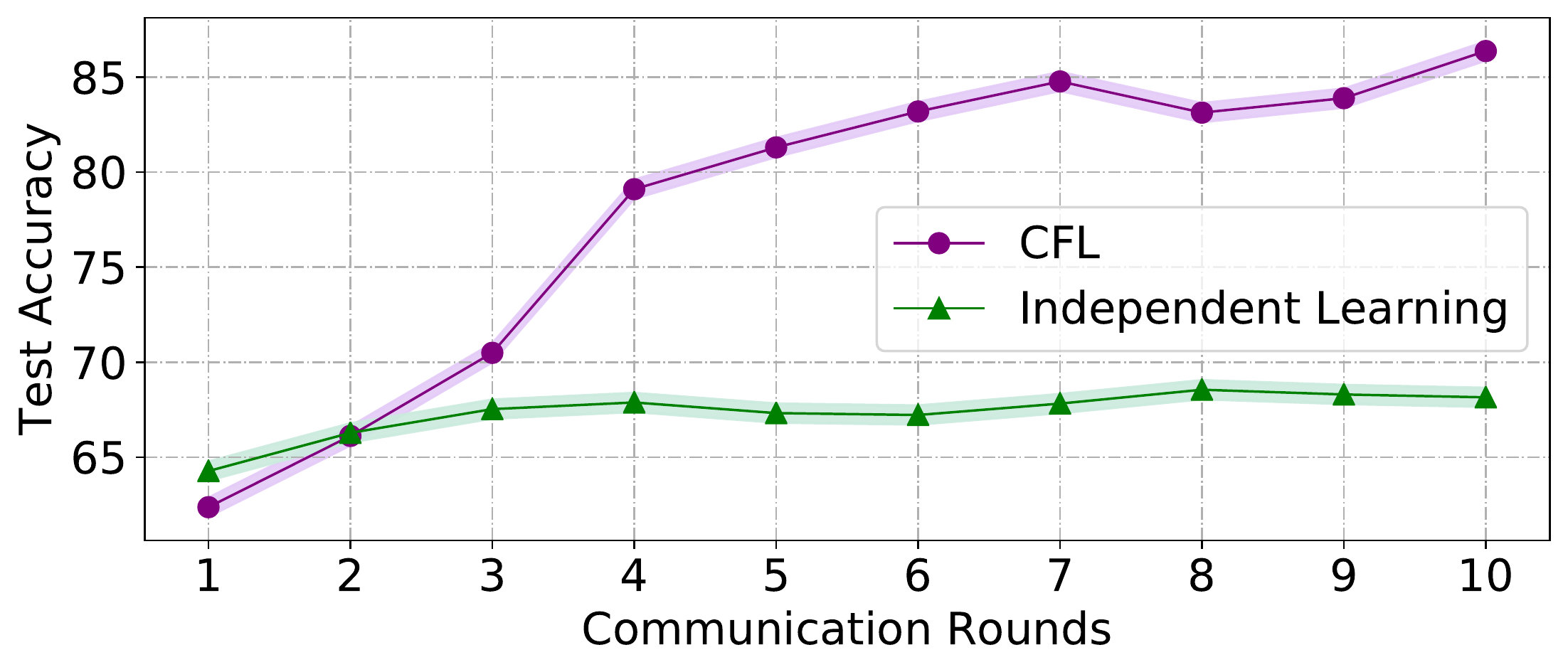} 
	}\\
	\subfigure[Distribution heterogeneity]{
	\includegraphics[width=0.36\textwidth]{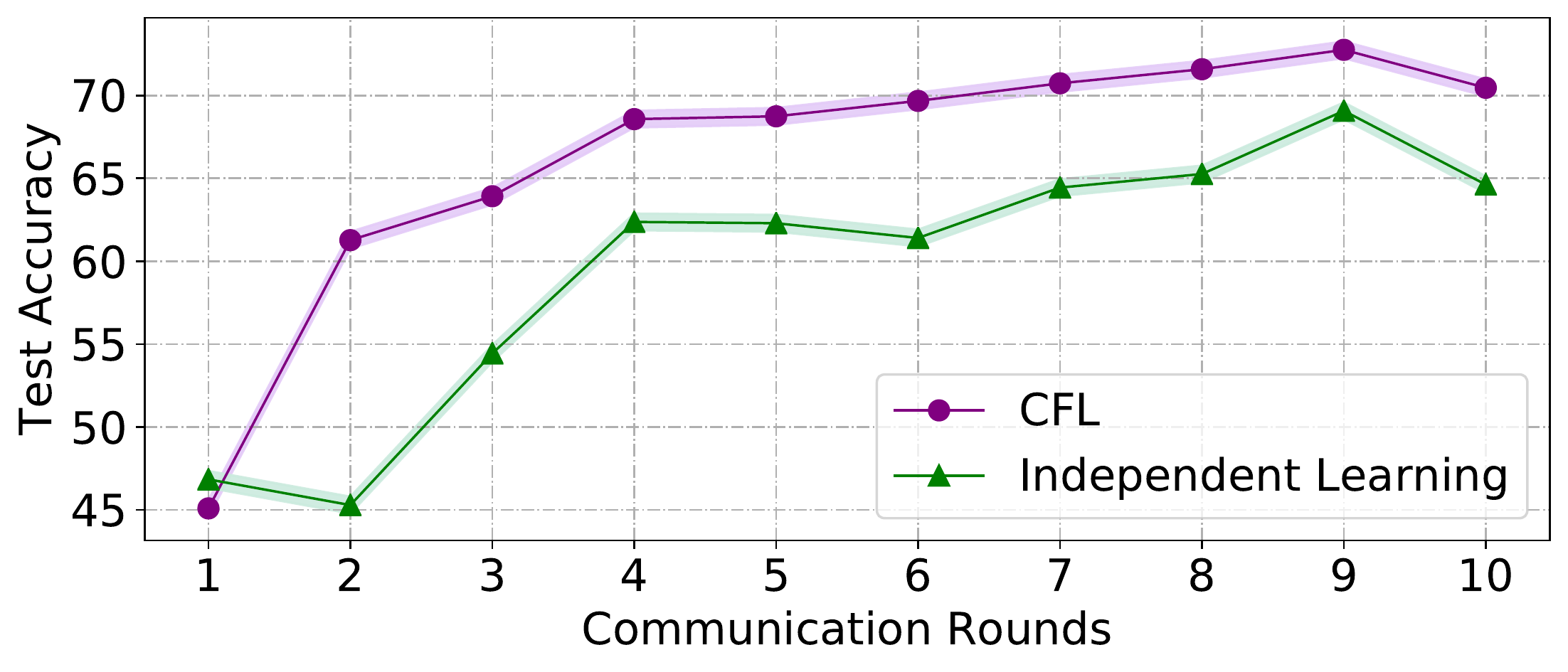} 
    }
	\vspace{-7pt}
	\caption{Comparison between CFL and independent learning with customized submodels. (a) Quality Heterogeneity: MNIST and 32 workers. (b) Distribution Heterogeneity: MNIST and 32 workers.}
	\label{noniidjoinindep}
\end{figure}
\begin{figure}[htp]
	\centering
	\subfigure[Worker 0 with gaussian fuzzy data]{
		\includegraphics[width=0.38\textwidth]{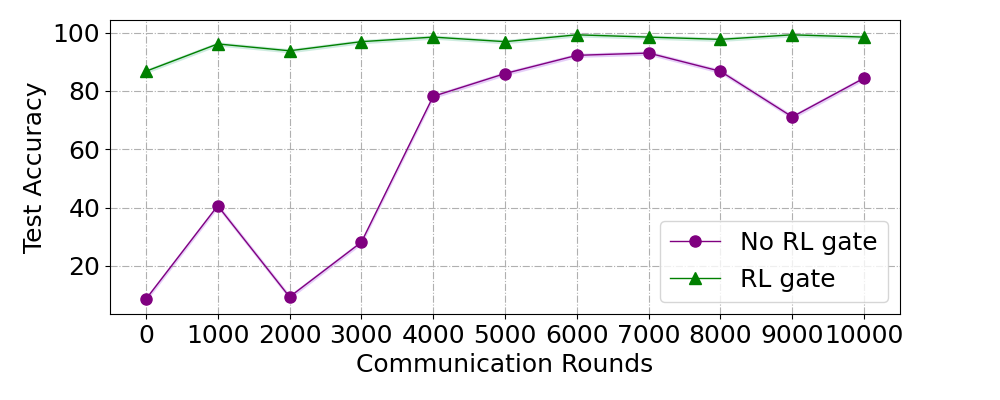} 		
	}\\
	\subfigure[Worker 1 with unprocessed standard data]{
		\includegraphics[width=0.38\textwidth]{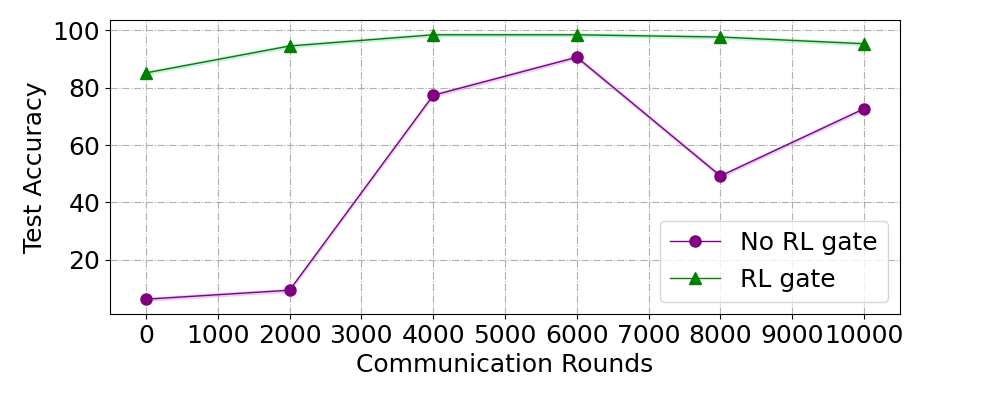} 
	}\\
	\subfigure[Worker 2 with sharpening data]{
		\includegraphics[width=0.38\textwidth]{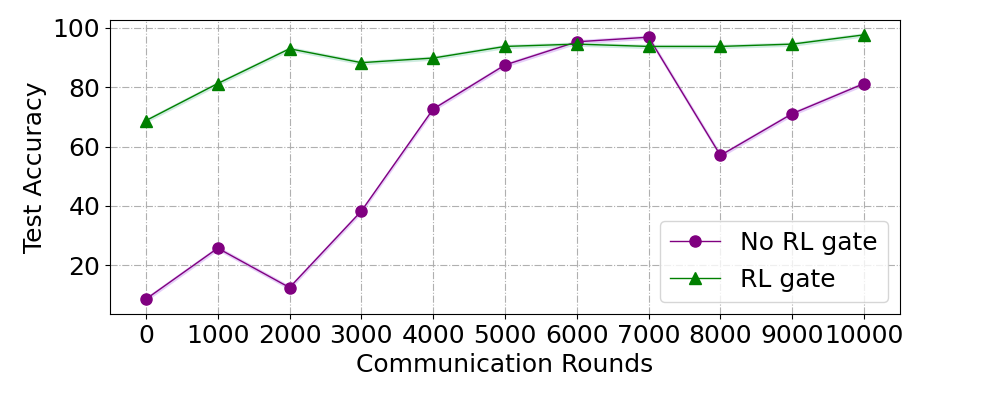} 
	}\\
	\subfigure[Computation Percentage]{
		\includegraphics[width=0.38\textwidth]{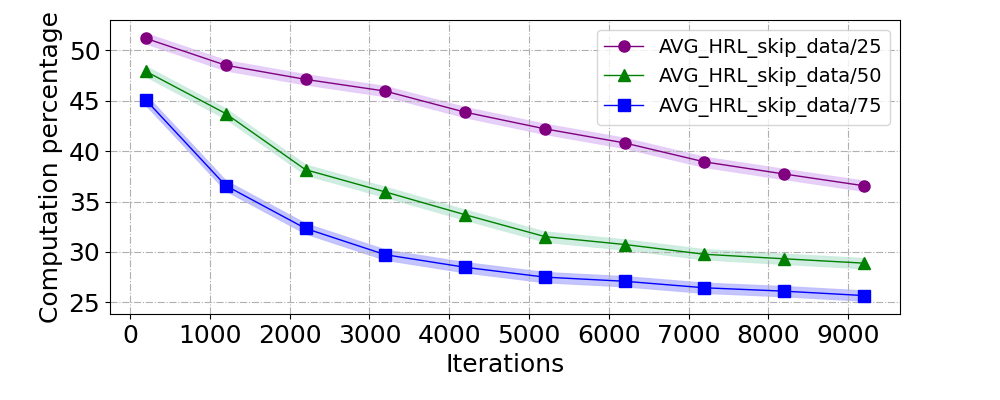} 
	}
	\caption{Comparison of accuracy and computational cost between the FL model using data quality-aware (with RL gate) and the common FL model without awareness under different data quality.}
	\label{gate}
\end{figure}
\textbf{} (\romannumeral1)
\textit{Heterogeneous Data Quality:} We generate Gaussian fuzzy data with three fuzzy degrees, unprocessed data, and sharpening data for all workers (randomly assigned), and then conduct the CFL and independent learning respectively. The results are shown in Fig~\ref{noniidjoinindep}(a). It is verified that the final test accuracy of the customized models in CFL is obviously higher than that of the independent learning method. This is because in CFL workers can learn from the experiences of others, which is absent in independent training. In reality, different edge devices often have heterogeneous data. 

\textbf{} (\romannumeral2)
\textit{Heterogeneous Data Distribution:} Fig~\ref{noniidjoinindep} (b) shows that the final test accuracy of CFL is higher than that of independent learning. Because in CFL, workers can learn and improve the local model from the parameter aggregation operation, while it is not possible for independent training. 

\subsection{Data quality-aware Parent Model}\label{subsec:par}
In this section, we show how the RL gates can benefit the parent model. We set different data qualities for different workers. After deploying RL gates on each layer of the original parent model, we first train it in advance on the server using a small public dataset with uniformly distributed categories and the worst data quality. This pre-trained model can then be used for submodel selection initially. The test accuracy of customized models is shown in Fig.~\ref{gate}(a-c). The results show that the RL gate-enabled submodel selection not only consumes less time to converge but also reduces both the training and inference time since the computation of some layers of customized models is waived. To better demonstrate the computation overhead reduction, the computational percentage curves in the training phase of the customized models for three workers are given in Fig.~\ref{gate}(d). This percentage is defined as the ratio of the number of layers that are actually calculated to the number of all layers of the model.

\section{Conclusion}
In this paper, we introduce a novel customized federated learning framework, which first takes the multi-dimensional heterogeneity in federated learning. Specifically, we design a novel aggregation algorithm to reduce the calculation delay and accuracy difference between the cooperative FL devices. 
What's more, CFL uses a specially designed data quality-aware central model via RL gate to accelerate reasoning and improve robustness in the face of data quality changes. Extensive experiments have proved the effectiveness of CFL.

\bibliographystyle{ieeetr}
\bibliography{bib}
\end{document}